%% file: main.tex
\documentclass[11pt,a4paper,dvipsnames]{article}
\usepackage[hyperref]{eacl2021}
\usepackage{times}
\usepackage{latexsym}
\usepackage{amsmath}
\usepackage{amssymb}
\usepackage{tabularx}
\usepackage{booktabs}
\usepackage{ifthen}
\usepackage{xspace}
\usepackage{multirow}
\usepackage[ruled]{algorithm2e}
\usepackage{comment}
\usepackage{graphicx}
\usepackage{subcaption}
\usepackage{tabularx}
\usepackage{enumitem}
\usepackage{xcolor}
\usepackage{tabulary}

\usepackage{microtype}

\aclfinalcopy 

\title{Unsupervised Extractive Summarization using Pointwise Mutual Information}

\author{Vishakh Padmakumar\\
  New York University \\
  \texttt{vishakh@nyu.edu} \\\And
  He He \\
  New York University \\
  \texttt{hehe@cs.nyu.edu} \\}

\date{}

\input std-macros

\input macros

\begin{document}
\maketitle
\begin{abstract}
Unsupervised approaches to extractive summarization usually
rely on a notion of sentence importance defined by
    the semantic similarity between a sentence and the document.
    We propose new metrics of relevance and redundancy using pointwise mutual information (PMI) between sentences,
which can be easily computed by a pre-trained language model.
    Intuitively, a relevant sentence allows readers to infer the document content (high PMI with the document),
    and a redundant sentence can be inferred from the summary (high PMI with the summary).
We then develop a greedy sentence selection algorithm to maximize relevance and minimize redundancy of extracted sentences.
We show that our method outperforms similarity-based methods on datasets
in a range of domains including news, medical journal articles, and personal anecdotes.

\end{abstract}

\input intro

\input approach

\input experiments
\input results
\input related
\input conclusion

\bibliographystyle{acl_natbib}
\bibliography{acl2020,all}

\clearpage
\appendix
\input appendix

\end{document}

%% file: std-macros.tex
\newcommand\pc[1]{\ensuremath{\left\{ #1 \right\}}} 

\newcommand\eqdef{\ensuremath{\stackrel{\rm def}{=}}} 
\newcommand\refeqn[1]{(\ref{eqn:#1})}

\newcommand\refsec[1]{Section~\ref{sec:#1}}

\newcommand\reffig[1]{Figure~\ref{fig:#1}}

\newcommand\reftab[1]{Table~\ref{tab:#1}}

\newcommand\refalg[1]{Algorithm~\ref{alg:#1}}

\ifthenelse{\isundefined{\definition}}{\newtheorem{definition}{Definition}}{}
\ifthenelse{\isundefined{\assumption}}{\newtheorem{assumption}{Assumption}}{}
\ifthenelse{\isundefined{\hypothesis}}{\newtheorem{hypothesis}{Hypothesis}}{}
\ifthenelse{\isundefined{\proposition}}{\newtheorem{proposition}{Proposition}}{}
\ifthenelse{\isundefined{\theorem}}{\newtheorem{theorem}{Theorem}}{}
\ifthenelse{\isundefined{\lemma}}{\newtheorem{lemma}{Lemma}}{}
\ifthenelse{\isundefined{\corollary}}{\newtheorem{corollary}{Corollary}}{}
\ifthenelse{\isundefined{\alg}}{\newtheorem{alg}{Algorithm}}{}
\ifthenelse{\isundefined{\example}}{\newtheorem{example}{Example}}{}

%% file: macros.tex
\DeclareMathOperator*{\argmax}{argmax}

\newcommand\eg{e.g.\xspace}
\renewcommand\nl[1]{\textit{``#1''}}

\newcommand\plm{\ensuremath{p_{\text{LM}}}}

\newcolumntype{L}[1]{>{\raggedright\let\newline\\\arraybackslash\hspace{0pt}}m{#1}}
\newcolumntype{C}[1]{>{\centering\let\newline\\\arraybackslash\hspace{0pt}}m{#1}}
\newcolumntype{R}[1]{>{\raggedleft\let\newline\\\arraybackslash\hspace{0pt}}m{#1}}

%% file: intro.tex
\section{Introduction}
\label{sec:intro}

Modern neural network-based approaches to summarization require a large amount of document-summary pairs that are usually unavailable outside of the news domain.
For example, summarization datasets of personal narratives and office meetings contain only a few hundred examples \citep{ouyang2017crowd,carletta2005ami}.
In this work, we tackle the problem of unsupervised extractive summarization
which aims to select important sentences from the document.
While there exists extensive prior work \citep{radev2000centroid,mihalcea2004textrank,liu2019text,zheng2019sentence},
most approaches rely on the assumption that important sentences are \emph{similar} to other sentences in the document. 
However, it is unclear if similarity-based features lead to meaningful content selection \citep{kedzie2018content}.

Inspired by recent work on formalizing the notion of importance in summarization \citep{peyrard2019information}, 
we propose metrics for \emph{relevance} and \emph{redundancy} based on pointwise mutual information (PMI).
Intuitively, a relevant summary allows the reader to maximally infer the document content, 
and a summary has minimal redundancy if each sentence in it provides additional information.
Therefore, we measure the relevance of a summary by its PMI with the document.
High relevance means that the probability of the document increases conditioning on the summary.
Similarly, we measure redundancy by PMI of sentence pairs \emph{within the summary}.
A sentence is redundant if seeing other sentences significantly increases its probability.

Based on the new metrics, we design a simple sentence extraction algorithm.
We estimate the PMI of sentence pairs by a pre-trained 
language model fine-tuned on in-domain documents.
We then use a simple sequential sentence selection algorithm for extractive summarization,
which greedily maximizes relevance and minimizes redundancy.

Experimental results show that our algorithm outperforms similarity-based methods
across multiple domains,
including news, personal stories, and medical articles.\footnote{Our code and pretrained models are available at \url{https://github.com/vishakhpk/mi-unsup-summ}.}

%% file: approach.tex
\section{Relevance and Redundancy}
\label{sec:formalization}

We begin by formalizing relevance and redundancy for summarization.
Consider a document of $n$ sentences $D = \pc{d_1, \ldots, d_n}$
and a summary of $m$ sentences $S = \pc{s_1, \ldots, s_m}$.
We would like to measure the relevance of $S$ to $D$
and the redundancy of $S$.

\paragraph{Relevance.}
Relevance measures how well the summary condenses the original text such that
we can infer its key content.
Specifically, a summary sentence is relevant if observing it reduces our uncertainty about (unseen) sentences in the document.
For example, the summary may contain the main link in the thread of conversation in the document.
We thus quantify the relevance of a summary sentence $s$ to a document sentence $d$ by their PMI:
\begin{align}
    \label{eqn:rel-sent-sent}
    \text{Rel}(s, d) \eqdef \text{pmi}(s;d) , 
\end{align}
which measures the dependence between $s$ and $d$.
A positive score means that $s$ and $d$ are very likely to co-occur,
thus seeing one implies the other.
A zero score means that $s$ and $d$ are independent.
A negative score means that $s$ and $d$ are unlikely to co-occur,
\eg contradicting sentences,
thus such a summary sentence is discouraged.
We further define the relevance of a summary $S$ to the document $D$
by the sum of sentence-level relevance:
\begin{align}
    \label{eqn:rel-sent-doc}
    \text{Rel}(S, D) \eqdef \sum_{s\in S}\sum_{d\in D} \text{Rel}(s, d) .
\end{align}

\paragraph{Redundancy.}
Redundancy measures how much overlap exists among the summary sentences.
It is typically measured by the semantic similarity between two sentences.
However, even if two sentences express different meanings,
there is redundancy if one is entailed by the other.
For example, consider:
\begin{enumerate}[itemsep=2pt]
    \item \nl{Michelle, of South Shields, Tyneside, says she feels like a new woman after dropping from dress size 30 to size 12.}
    \item \nl{Michelle weighed 25st 3lbs when she joined the group in April 2013 and has since dropped to 12st 10lbs.}
\end{enumerate}
Though expressing different information, both imply Michelle's weight loss.

Given a summary sentence, we want to assign a score proportional to the amount of information in the sentence which is already present in the rest of the summary.
Therefore, we quantify the redundancy of a sentence $s$ given another sentence $s'$ by
their dependence in terms of PMI:
\begin{align}
    \label{eqn:red}
    \text{Red}(s, s') \eqdef \text{pmi}(s;s'). 
\end{align}
Similarly, the redundancy of a summary $S$ is defined by the total redundancy of all sentence pairs:
\begin{align}
    \text{Red}(S) \eqdef \sum_{i=1}^m\sum_{j=i+1}^m\text{pmi}(s_i;s_j) .
\end{align}

\paragraph{Estimate PMI.}
By definition, $\text{pmi}(s;d) = \log\frac{p(s\mid d)}{p(d)} = \log\frac{p(d\mid s)}{p(s)}$.
Since both $s$ and $d$ are sentences,
we use a language model, $\plm$,  to estimate
the probabilities.
Conditional probabilities are comoputed by considering the condition sentence as the prefix. 

Note that while PMI can be computed in two equivalent ways according to the definition,
the estimates from a language model do not guarantee that
$\frac{\plm(d\mid s)}{\plm(d)} = \frac{\plm(s\mid d)}{\plm(s)}$.
Thus we choose to condition on the summary sentence:
\begin{align}
\text{pmi}(s;d) \eqdef \log\frac{\plm(d\mid s)}{\plm(d)} .
\end{align}
This is consistent with our definition of relevance:
\emph{seeing the summary}, how well we can estimate the document content.
For redundancy, we condition on the earlier sentence:
\begin{align}
\text{pmi}(s_i;s_j)\eqdef
    \begin{cases}
        \log\frac{\plm(s_j\mid s_i)}{\plm(s_j)} & \text{if } i < j \\
        \text{pmi}(s_j;s_i) & \text{otherwise}
    \end{cases}
    ,
\end{align}
since a sentence is redundant if it can be inferred from previous sentences.

\section{Sequential Sentence Extraction}
\label{sec:interpolated}

Given relevance and redundancy defined above,
we aim to select important sentences from the document
that maximize relevance and minimizes redundancy.
We consider a weighted combination of the two criteria:
\begin{align}
    \label{eqn:objective}
    \max_{S\subseteq D}\; \lambda_1 \text{Rel}(S, D) + \lambda_2\text{Red}(S)
    \; \text{s.t.}\: |S| \le k ,
\end{align}
where $|S|$ denotes the number of sentences in the summary.
This is a combinatorial problem that is expensive to solve  when $k$ is large.
Therefore, we solve it approximately by selecting sentences sequentially in a greedy fashion.
Given the previously selected sentences,
we select the next sentence from the document that maximally improves the objective \refeqn{objective}
until $k$ sentences are selected.
Our full algorithm is shown in \refalg{sentence_selection},
where $\Delta(s)$ denotes the change incurred by the new summary $S\cup\pc{s}$.

\begin{algorithm}[h]
\DontPrintSemicolon
{
    $S := \emptyset$\;
    \For{$j\gets1$ \KwTo $k$}{
        \For{$s \in D\setminus S$}{
            $\Delta(s) := \lambda_1 \text{Rel}(\pc{s}, D) + \lambda_2\sum_{s'\in S}\text{Red}(s', s)$
        }
        $s_j := \argmax_{s\in D\setminus S}\Delta(s)$\;
        $S := S \cup \pc{s_j}$\;
    }
    \Return $S$\;
}
\caption{ExtractSentences($D$, $k$)}
\label{alg:sentence_selection}
\end{algorithm}

%% file: experiments.tex
\section{Experiments}
\label{sec:experiments}

\paragraph{Datasets.}
We evaluate on a diverse range of domains:
(i) news articles: CNN-Dailymail (CNN-DM) \cite{see2017point} and XSum \cite{narayan2018details};
(ii) personal anecdotes: Reddit \cite{ouyang2017crowd} and Reddit-TIFU \cite{kim2019abstractive};
(iii) scientific articles: PubMed \cite{kedzie2018content}.
Further, these datasets allow us to evaluate on highly abstractive summarization (XSum), small data (Reddit), and long documents (PubMed).
Details of the datasets are shown in Appendix \ref{app:datasets}.

\paragraph{Baselines.}
We compare our approach against the following unsupervised extraction methods:
(i) \textbf{heuristic}: lead-$k$ which selects the first $k$ sentences.
(ii) \textbf{similarity-based}:
TextRank \cite{barrios2016variations, mihalcea2004textrank} and PacSum \cite{zheng2019sentence}
which use graph-based selection
with similarity metrics based on tf-idf sentence vectors and BERT embeddings, respectively.\footnote{
For TextRank, \citet{zheng2019sentence} showed that tf-idf works better than BERT embeddings so we follow the same.
}

To ablate the contribution of PMI, we also include a variant of our algorithm which uses cosine similarity of tf-idf sentence representations to measure redundancy and relevance of sentence pairs. 
Additionally, we include two \textbf{reference} methods:
oracle extraction\footnote{
We select sentences greedily to optimize the Rouge-1 score against the reference \cite{nallapati2017summarunner}. 
}
and the state-of-the-art supervised approaches:
For CNN-Dailymail, XSum, Reddit-TIFU and Pubmed the results are from \citet{zhang2019pegasus} and for Reddit from \citet{ouyangphdthesis}. 

\paragraph{Language model fine-tuning.}
For our method, we use the pre-trained GPT-2 large model \cite{radford2019language} to calculate the PMI.
To adapt the language model to specific domains, we fine-tune it on the training documents (excluding the gold summaries) in each dataset.
To make it better fit our task of 
estimating the probability of one sentence given another, as described in \refsec{formalization}, we fine-tune GPT-2 on two-sentence segments (as opposed to a long stream of tokens).
Each segments consist of a pair of sentences from the document.\footnote{Our preliminary results show that this works slightly better than standard LM fine-tuning.}

\paragraph{Implementation details.}
We preprocess the documents with spaCy \cite{honnibal2017spacy2} to split the text into sentences.
All hyperparameters are tuned on 
200 randomly sampled document-summary pairs selected from the validation set
to optimize the Rouge-1 F-measure,
including $\lambda_1$ and $\lambda_2$ in our method which balances relevance and redundancy scores in Equation \refeqn{objective},
the number of keywords in TextRank and the number of sentences to select for all extractive methods.

To select the values of $\lambda_1$ and $\lambda_2$ we 
run a grid search at intervals of 0.1 from -2 to 2 for both. For all datasets, the best weighting was 2 for relevance and -2 for redundancy.\footnote{
    Note that there are multiple optimal values, e.g. 1 and -1, the selected values are based on tie-breaking in our algorithm. 
}
The weights are intuitive because we want to maximize relevance ($\lambda_1$) and minimize redundancy ($\lambda_2$).
For the Lead-$k$ baselines, $k$ was 3 for CNN-Dailymail, Reddit-TIFU and XSum, 4 for Reddit and 9 for PubMed.

%% file: results.tex
\subsection{Results}
\renewcommand{\tabcolsep}{2.5pt}
\begin{table*}[h!]
    \centering
    \footnotesize
        \begin{tabular}{|c|rrr|rrr|rrr|rrr|rrr|}
        \hline
        \multirow{2}{*}{Dataset} & \multicolumn{3}{c|}{CNN-DM} & \multicolumn{3}{c|}{XSum} & \multicolumn{3}{c|}{Reddit} &  \multicolumn{3}{c|}{Reddit-TIFU} &
        \multicolumn{3}{c|}{PubMed}\\
        & R-1 & R-2 & R-L & R-1 & R-2 & R-L & R-1 & R-2 & R-L & R-1 & R-2 & R-L & R-1 & R-2 & R-L \\
        \hline
            Lead-k & 39.69 & 17.22 & 24.82 & \textbf{19.48} & 2.62 & 12.49 & 26.52 & 7.05 & 18.80 & 15.4 &    2.24 &  10.81 & 37.34 & 10.54 & 18.00 \\
            TextRank & 34.11 & 12.78 & 22.51 & 19.04 & 3.05 & \textbf{12.64} & 23.76 & 7.90 & 16.41 & 18.70 & 3.49 & 13.08 & 46.73 & 17.28 & 22.28 \\
        PacSum & \textbf{40.26} & \textbf{17.55} & \textbf{24.92} & 19.44 & 2.71 & 12.44 & - & - & - & - & - & - & - & - & - \\
        \hline
        Ours (tf-idf) & 34.17 & 12.99 & 22.44 & 18.14 & 3.14 & 12.31 & 24.14 & 8.19 & 16.96 & 18.24 & 3.18 &  12.45 & 46.54 & 17.77 & 22.66 \\
            Ours (PMI) & 36.68 & 14.52 & 23.32 & 19.07 & \textbf{3.22} & 12.47 & \textbf{28.18} & \textbf{8.60} & \textbf{20.26} & \textbf{18.93} & \textbf{3.72} & \textbf{13.11} & \textbf{46.84} & \textbf{17.81} & \textbf{23.01} \\
        \hline
        \hline
        Oracle Ext & 53.53 & 29.56 & 37.21 & 31.85 & 7.51 & 20.87 & 36.63 & 15.24 & 27.87 & 34.96 & 10.75 & 24.91 & 54.88 & 23.72 & 28.30 \\
        Supervised & 44.17 & 21.47 & 41.11 & 47.21 & 24.56 & 39.25 & 20.90 & 5.40 & 18.90 & 26.63 & 9.01 & 21.6 & 45.09 & 19.56 & 27.42 \\
        \hline
    \end{tabular}
    \caption{Rouge-1/2/L F-Measure scores of unsupervised extractive methods, oracle extraction, and SOTA results using surpervised learning on CNN-DM, XSum, Reddit, Reddit-TIFU and PubMed.
    The best results among unsupervised methods are in bold.
    Our PMI-based extractor outperforms similarity-based methods on non-news domains,
    and is comparable on news domains.
    (PacSum results on non-news domains are not reported as the released model is fine-tuned only on the news domain.) }
    \label{tab:rouge}
\end{table*}

\begin{table*}[h!]
    \centering
    \footnotesize
    \begin{tabular}{|m{0.3\textwidth}|m{0.3\textwidth}|m{0.3\textwidth}|}
        \hline
        Gold Summary & Ours(PMI) & PacSum \\
        \hline
        Cillian McCann was filmed by his mother Toni at seven weeks old. In the clip, the little boy can clearly be seen trying to speak to his family. After several attempts he manages to say "hello". The average child can say six words by the time they reach 18 months. & Whose adorable son Cillian said his first word at just seven weeks old. \textbf{In the video Cillian is seen struggling to get his word out, but with a bit of encouragement from his mother he finally says hello.} Toni says that Cillian was very alert from a young age and had been trying to make out words since he was just five weeks old . & Most parenting advice says you don't have to worry if your baby doesn't start speaking until around 18 months. The tiny tot, who is now nine weeks old, was filmed by his 36-year-old mother who says that she knew he had been trying to communicate for a while. \textbf{Cillian has three older sisters, Toni revealed that her little girls, Sophie(bottom right), Eva(bottom left) and Ellie(top), did not start talking at such an early age.}\\
        \hline
    \end{tabular}
    \caption{Example of summaries selected by PMI vs PacSum. PacSum selected the sentence (bolded) about the child's siblings.
    It mentions talking at an early age which is the main theme of the article.
    However, it does not inform us how the child started talking in the video. 
    Our method using PMI selected sentences focusing on the content of the video and the child.}
    \label{tab:example}
\end{table*}

We evaluate all methods on the five datasets using the  Rouge-1/2/L \cite{lin2004rouge} F-measure.
Table \reftab{rouge} shows our main results. 

\paragraph{PMI vs similarity.}
We first compare PMI and tf-idf in our framework.
The results  (Ours (PMI) vs Ours (tf-idf))
show that measuring relevance and redundancy
using PMI is better than word overlap,
especially on narratives (Reddit).
The Reddit writing style is less reporting facts like in news and more describing a sequence of events,
thus it is helpful to capture the dependence between events.
We show an example contrasting the two metrics in \reftab{example}.
Further, our method achieves better or comparable results across all datasets compared to other similarity-based methods. 
The oracle extraction is predictably the best result across all datasets. Our extractive approach outperforms the supervised baseline on the smallest dataset, Reddit, demonstrating the utility of unsupervised approaches in this setting.
More examples are shown in Appendix \ref{app:examples}.

    \paragraph{Extraction algorithm.}
Another important component in extractive methods is the sentence selection/decoding algorithm.
The most common approach is to select sentence greedily according to certain objective.
TextRank uses a graph-based method inspired by PageRank.
    However, in \reftab{rouge} we see that TextRank and Ours (tf-idf) (using greedy selection)
achieve similar results,
showing that the selection algorithm does not have a large impact,
which is also found by \citet{zheng2019sentence}.

\paragraph{Ablation study.}
To understand the contribution of relevance and redundancy in the proposed metric,
we conducted an ablation study on CNN/DM and Reddit-TIFU.
In \reftab{ablation}, we see that relevance alone does well, but augmenting it with redundancy obtains the best performance across all metrics.
Minimizing redundancy alone works poorly because it cannot identify important content. 

\renewcommand{\tabcolsep}{2pt}
\begin{table}[ht]
    \centering
    \footnotesize
    \begin{tabular}{|>{\centering\arraybackslash}p{2cm}|rrr|rrr|}
        \hline
        \multirow{2}{*}{Method} & \multicolumn{3}{c|}{CNN-DM} & \multicolumn{3}{c|}{Reddit-TIFU} \\
         & R-1 & R-2 & R-3 & R-1 & R-2 & R-3 \\
        \hline
        Only Rel. & 35.17 & 13.79 & 22.91 & 18.45 & 3.67 & 12.90 \\
        \hline
        Only Red. & 23.85 & 6.47 & 15.48 & 16.41 & 2.35 & 11.66 \\
        \hline
        Rel. and Red. & \textbf{36.68} & \textbf{14.52} & \textbf{23.32} & \textbf{18.93} & \textbf{3.72} & \textbf{13.11} \\
        \hline
    \end{tabular}
    \caption{Ablation Results on CNN-DM and Reddit-TIFU.  We observe that the combination of relevance and redundancy yields the best performance across all metrics.}
    \label{tab:ablation}
\end{table}

\begin{figure}[ht!]
    \centering
    \includegraphics[width=0.5\textwidth, keepaspectratio]{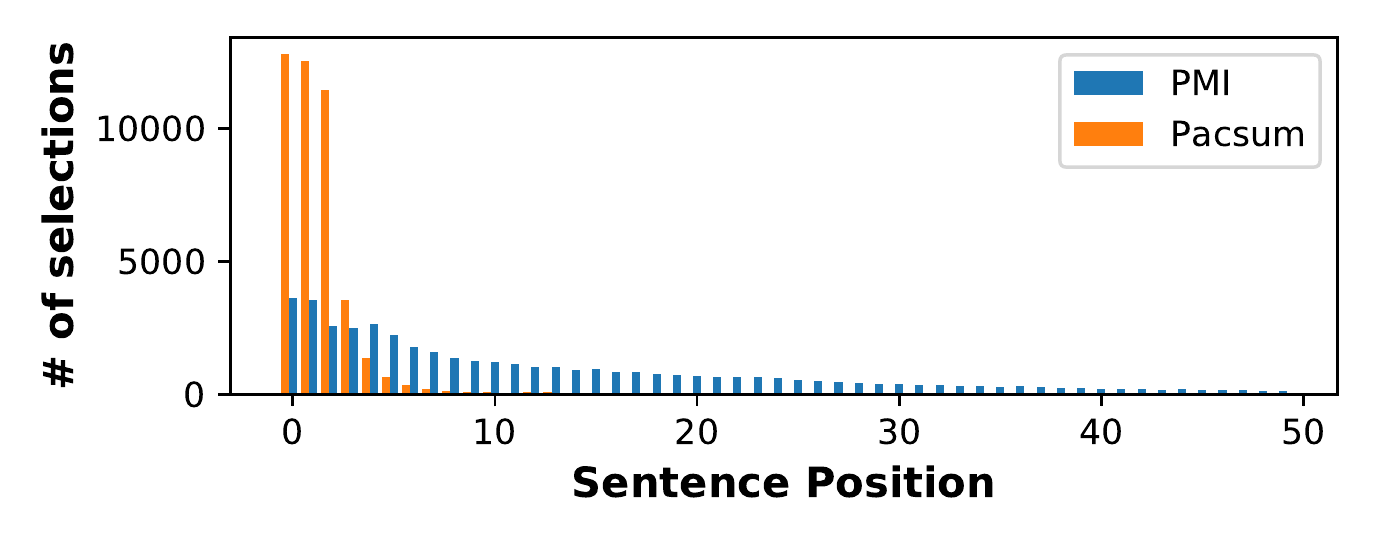}
    \caption{
        Histogram of the position of sentences selected by our method and PacSum on CNN/DM.
        PacSum uses position information which allows it to take advantage of the lead bias.
        In contrast, our method is position-agnostic but still captures the fact that
        earlier sentences are more important in news articles.
    }
    \label{fig:distribution}
\end{figure}

\paragraph{Position bias in PacSum.}
\label{app:position}
One may wonder why PacSum and lead-$k$ significantly outperform other extractive methods on CNN-DM.
We hypothesize that they take advantage of the lead bias on CNN/DM. 
\reffig{distribution} shows histograms of the positions of summary sentences selected by our method and PacSum on CNN/DM.
Notably, 82.3\% of sentences selected by PacSum were in the first three, 
and this value drops to 21.4\% in our method.
This provides an empirical explanation as to why PacSum is so far ahead of the other extractive approaches on CNN-DM. The authors \cite{zheng2019sentence} also noted a drop in performance when positional information was removed in PacSum. In addition, their performance degrades on XSum (which doesn't suffer from lead bias). 
A concurrent work \cite{xu2020unsupervised} performed a similar analysis, observing the reliance on position information of PacSum in the news domain.

%% file: related.tex
\section{Related Work}
\label{sec:related}
\paragraph{Similarity-based summarization.} Most unsupervised extractive summarization
methods
rely on sentence pair similarity as a proxy for importance \cite{zheng2019sentence, mihalcea2004textrank, barrios2016variations,erkan2004lexrank}
and use variants of the Pagerank \cite{page1999pagerank} algorithm to perform selection. We propose PMI as an alternative and compare it to a concurrent similarity based approach.

\paragraph{Leveraging pretrained language models.}
Scoring a sentence on its ability to predict subsequent sentences using a language model has been adapted for sentence summarization. 
\cite{west2019bottlesum}. \citet{zhou2019simple} use two language models, a generic pretrained language model for contextual matching and a task specific one to enforce fluency. A concurrent work \cite{xu2020unsupervised} used sentence level transformer self-attentions and probabilities to rank sentences for unsupervised extractive summarization.  We use the language model to compute PMI, which then scores sentences on relevance and redundancy as criteria for selection. 

\paragraph{Diversity in content selection.} 
Maximal Marginal Relevance \cite{goldstein1998summarization} has been used to produce summaries that prioritize diversity in selected content. 
A similarity metric is used to produce summaries based on similarity to a query while maintaining diversity among selected sentences in various domains \cite{chandu2017tackling}. This can be seen as analogous to our comparison to our approach tf-idf
based selection.

%% file: conclusion.tex
\section{Conclusion and Discussion}
\label{sec:conclusion}
We propose metrics for relevance and redundancy in summarization based on pointwise mutual information,
and an unsupervised extractive summarization algorithm using pre-trained language models.
We demonstrate the effectiveness of our method on both news and non-news domains.
Supervised models often learn the lead bias in the datasets and degrade significantly when such hues are absent \cite{kedzie2018content}.
Furthermore, even human evaluation of content selection has large variance \cite{nenkova2007pyramid,chaganty2018evaluation}.
Our work is a first step towards formalizing a notion of importance that informs algorithm design in summarization.
We believe it is important to have a better formalization of content importance in terms of both task definition/evaluation and modeling. 
We hope our results will spur more work in this direction.

\section*{Acknowledgement}
This work was partly supported by Samsung Advanced Institute of Technology (Next Generation Deep Learning: From Pattern Recognition to AI) and Samsung Research (Improving Deep Learning Using Latent Structure).

%% file: appendix.tex
\section{Dataset Details}
\label{app:datasets}
CNN/DM is known to have a very strong extractive Lead-3 baseline as is common in the news domain. XSum contains summaries of BBC news articles but is highly abstractive in nature. The Reddit dataset is a small corpus of around 500 personal stories shared on Reddit with abstractive and extractive summaries. For Reddit-TIFU, we use the TIFU-long subset as used in \citet{zhang2019pegasus}. The Reddit-TIFU didn't come with a train split and since we look at unsupervised methods, we used 200 pairs as validation data to decide parameters and report test results on the rest.
The PubMed dataset contains longer medical journal articles with the corresponding abstracts functioning as the groundtruth summaries.

\begin{table}[ht!]
    \centering
        \begin{tabular}{|c|c|c|c|}
            \hline
            \textbf{Dataset} & \textbf{Train} & \textbf{Validation} & \textbf{Test} \\ 
            \hline
            CNN-DM & 287,113 & 13,368 & 11,490 \\  
            \hline
            XSum & 204,045 & 11,332 & 11,334 \\
            \hline 
            Reddit  & 404 & 24 & 48 \\
            \hline
            Reddit-TIFU & - & 200 & 41139 \\
            \hline
            PubMed & 21,250 & 1,250 & 2,500\\
            \hline
        \end{tabular}
    \caption{Size of train, validation and test splits of datasets used}
    \label{tab:datasets}
\end{table}

\section{Analysis of Examples}
\label{app:examples}
Table \ref{tab:examples} shows some example summaries from the CNN-Dailymail validation set in comparison to extractive candidate summaries obtained for the correponding documents using the baseline Lead-3 approach, our Interpolated PMI based approach and the PacSum approach \cite{zheng2019sentence} that uses sentence similarity to obtain state-of-the-art Rouge results on the dataset. These are shown to highlight the difference between using PMI and similarity for sentence selection. In the first example, the gold summary details about how medical information regarding two patients was leaked to sales representatives. The similarity based approach selects all three sentences associated with only one of the patients whereas the PMI based approach yields a summary that contains information about both. Once the first sentence concerning the first patient is selected, all sentences associated with it are penalised by a corresponding amount resulting in a more well rounded selection of information. Similarly, in the second example the sentence that details how the child overcame his initial struggles to speak after some encouragement from his parent was only selected by the PMI approach. This summarises exactly what happens in the video clip being spoken about and the same point is highlighted even in the gold summary. The contents of the paragraph can be easily understood when the information about the clip is used as a context. The rest of the paragraph talks goes into how the child spoke faster than his siblings which explains the selection made by PacSum. The third example highlights the issue of PacSum being identical to the Lead-3 baseline by modelling position information present in the dataset. In the fourth example, the PacSum based approach selects two sentences with quotes that have negative connotations while the one selected by PMI about how the protagonist could not forgive himself could serve to better explain the need for an intervention on the Dr. Phil show.\\

The purpose of this is to highlight that the intangible nature of the definition of relevance. The content selected varies between PMI and sentence similarity and each might find an application in the right setting. It again highlights the need to consider what one expects from the summarisation task.

\begin{table*}[ht]
    \tiny
    \centering
    \begin{tabular}{ | m{3.6cm} | m{3.6cm}| m{3.6cm} | m{3.6cm} | } 
        \hline
        \textbf{Gold} & \textbf{Interpolated PMI} & \textbf{PacSum} & \textbf{Lead3} \\ 
        \hline
        Tim Esworthy, 66, has a prosthetic limb after losing his leg in a workplace incident and said he had been targeted by cold callers selling products easing joint pain. Christine Lewis, 62, is wheelchair-bound following a brain haemorrhage. She is also on list of people obtained by the mail and has been targeted by stairlift salesmen.  & Tim Esworthy, 66, from colchester, was 'absolutely appalled' to find his private medical details had been sold. Christine Lewis, who is recovering from a brain haemorrhage she had 12 years ago, was on a list of people who have mobility problems obtained by the mail and has been targeted by stairlift salesmen cold calling her. 'They shouldn't have my information, especially if they know I'm disabled because they are targeting me because they think I'm vulnerable.' &  Tim Esworthy, 66, from colchester, was 'absolutely appalled' to find his private medical details had been sold. Retired financial services manager Tim Esworthy was 'absolutely appalled' to find his private medical details had been sold . They know they can target vulnerable people because they have their medical information. & Tim Esworthy, 66, from Colchester, was 'absolutely appalled' to find his private medical details had been sold. Case 1 : Pensioner who lost leg at work. Retired financial services manager Tim Esworthy was 'absolutely appalled' to find his private medical details had been sold.\\
        \hline
        Cillian McCann was filmed by his mother Toni at seven weeks old. In the clip, the little boy can clearly be seen trying to speak to his family. After several attempts he manages to say "hello". The average child can say six words by the time they reach 18 months. & Whose adorable son Cillian said his first word at just seven weeks old. In the video Cillian is seen struggling to get his word out, but with a bit of encouragement from his mother he finally says hello. Toni says that Cillian was very alert from a young age and had been trying to make out words since he was just five weeks old . & Most parenting advice says you don't have to worry if your baby doesn't start speaking until around 18 months. The tiny tot, who is now nine weeks old, was filmed by his 36-year-old mother who says that she knew he had been trying to communicate for a while. Cillian has three older sisters, Toni revealed that her little girls, Sophie(bottom right), Eva(bottom left) and Ellie(top), did not start talking at such an early age. & Most parenting advice says you don't have to worry if your baby doesn't start speaking until around 18 months. Whose adorable son Cillian said his first word at just seven weeks old. The tiny tot, who is now nine weeks old, was filmed by his 36-year-old mother who says that she knew he had been trying to communicate for a while.\\ 
        \hline
        Ashleigh humphrys, 20, died in a hit-and-run early on sunday morning. Police believe the driver of the car was heading to work. A man is assisting police with their investigations after the death. Ms Humphrys was walking home after celebrating her birthday with friends. A security guard rang police after she was walking disorientated. CCTV footage shows two taxis stop near her before she was struck and put hazard lights on. Then a car drove past the taxis, mounted the footpath before swerving back onto the road and driving off. A taxi is said to have been seized and police are talking to a person 'within the vicinity' at the time of the incident. & Brisbane woman Ashleigh Humphrys died in a hit-and-run incident after deciding to walk from Toowong to her Seventeen Mile Rocks home in Brisbane after having an argument with a friend while they were out celebrating her 20th birthday. Only moments later the guard, who was still on the phone to police while driving around trying to find Ms Humphrys, discovered her dead on the road at the city end of the western freeway. Just before Ms Humphrys was hit, CCTV footage shows two taxis stop near the woman and put their hazard lights on before a car drove past the taxis, mounted the footpath and then swerved back onto the road before driving off. &  The driver of a car that hit and killed a young woman in the early hours on Sunday morning was on the way to work, police believe. Brisbane woman Ashleigh Humphrys died in a hit-and-run incident after deciding to walk from Toowong to her Seventeen Mile Rocks home in Brisbane after having an argument with a friend while they were out celebrating her 20th birthday. Now, after it was revealed that a man was assisting police with their investigations, officers have said they believe he was on his way to work and went to his shift as normal on sunday, the Courier Mail reported. & The driver of a car that hit and killed a young woman in the early hours on Sunday morning was on the way to work, police believe. Brisbane woman Ashleigh Humphrys died in a hit-and-run incident after deciding to walk from Toowong to her Seventeen Mile Rocks home in Brisbane after having an argument with a friend while they were out celebrating her 20th birthday. Now, after it was revealed that a man was assisting police with their investigations, officers have said they believe he was on his way to work and went to his shift as normal on sunday, the Courier Mail reported.\\
        \hline
         Dr. Phil Mcgraw staged a highly-charged intervention with Nick Gordon last Thursday. With his mother, Michelle, by his side a sobbing Gordon talked about missing Bobbi Kristina. Gordon is now in rehab after the intervention having been drinking heavily and taking xanax. Girlfriend Bobbi Kristina has been in a medically induced coma since January 31 and Gordon has not been allowed to see her. The dramatic intervention will air Wednesday on the Dr Phil show. & Amid scenes of high emotion, an often incoherent Gordon admitted drinking heavily and taking xanax, for which he has a prescription, in an attempt to deal with life since Bobbi Kristina was found face down and unresponsive in her bathtub on January 31. Breakdown: With his mother, Michelle, by his side Nick Gordon struggles to stay coherent as he is questioned by Dr Phil. According to his mother, Michelle, Gordon can not forgive himself for his 'failure' to revive Bobbi Kristina &  Weeping and wailing Nick Gordon, the troubled fiancé of Bobbi Kristina Brown, has admitted that he has twice tried to kill himself and confessed: "I'm so sorry for everything." Asked if he still intended to kill himself he said: "If anything happens to Krissi I will." Amid scenes of high emotion, an often incoherent Gordon admitted drinking heavily and taking xanax, for which he has a prescription, in an attempt to deal with life since Bobbi Kristina was found face down and unresponsive in her bathtub on January 31. & Weeping and wailing Nick Gordon, the troubled fiancé of Bobbi Kristina Brown, has admitted that he has twice tried to kill himself and confessed : "I'm so sorry for everything." Gordon, 25, was speaking to Dr Phil Mcgraw in a dramatic intervention due to air on Wednesday, Daily Mail online can reveal. Asked if he still intended to kill himself he said: "If anything happens to Krissi I will." \\
         \hline
    \end{tabular}
    \caption{Example summaries obtained from the CNN-Dailymail validation set compared to the corresponding extractive candidate summary obtained using Interpolated PMI, PacSum (State-of-the-art unsupervised summary using sentence similarity) and the Lead-3 Baseline}
    \label{tab:examples}
\end{table*}
\begin{center}

\end{center}